# Keystroke Dynamics Authentication For Collaborative Systems

Romain Giot, Mohamad El-Abed, Christophe Rosenberger
*GREYC Laboratory , ENSICAEN – University of CAEN – CNRS*
*romain.giot@ensicaen.fr, melabed@greyc.ensicaen.fr,christophe.rosenberger@greyc.ensicaen.fr*

## ABSTRACT

*We present in this paper a study on the ability and the benefits of using a keystroke dynamics authentication method for collaborative systems. Authentication is a challenging issue in order to guarantee the security of use of collaborative systems during the access control step. Many solutions exist in the state of the art such as the use of one time passwords or smart-cards. We focus in this paper on biometric based solutions that do not necessitate any additional sensor. Keystroke dynamics is an interesting solution as it uses only the keyboard and is invisible for users. Many methods have been published in this field. We make a comparative study of many of them considering the operational constraints of use for collaborative systems.*

## 1. INTRODUCTION

Collaborative systems are useful for engineers and researchers to facilitate their work [1,2]. User authentication on these systems is important as it guarantees that somebody works or shares data with the right person and that only authorized individuals can access to a collaborative application or data.

In security systems, *authentication* and *authorization* are two complementary mechanisms for determining who can access information resources over a network. *Authentication* systems provides answers to two questions: (i) who is the user? and (ii) is the user really who he/she claims himself to be? *Authorization* is the process of giving individuals access to system objects based on their identity. Authorization systems provide the answers of three questions: (i) is user X authorized to access resource R?, (ii) is user X authorized to perform operation P? and (iii) is user X authorized to perform operation P on resource R?

The authentication process can be based on a combination of one or more authentication factors. The four widely recognized factors to authenticate humans to computers are:
- Something the user **knows**, as a password, a pass-phrase or a PIN code.
- Something the user **owns**: a security token, a phone or a SIM card, a software token, a navigator cookie.
- Something that **qualifies the user:** a fingerprint, a DNA fragment, or a voice pattern for example, or something that qualifies **the behavior of a user:** signature dynamics, keystroke dynamics or a gait for example.

The well-known ID/password (static authentication) is far the most used authentication method. It is widely used despite its obvious lack of security. This fact is due to the ease of implementation of this solution, and to the instantaneous recognition of that system by the users that facilitates its deployment and acceptance.

Static authentication suffers of many drawbacks, as it is a low-security solution when used without any precaution:
- Passwords can be shared between users.
- Password can be stolen on the line by an eavesdropper, and used in order to gain access to the system (replay attack).

- Physical attacks can easily be done, by a camera recording a PIN sequence as it is typed, or by a key logger (either software or hardware) to record a password on a computer.
- Passwords are also vulnerable to guessing attacks, like brute force attacks or dictionary attacks or through social engineering attacks (highly facilitate by social networks providing many personal information of an individual). However, it is possible to limit the scope of these attacks by adding a delay (of about one second) before entering the password and after entering a wrong password.

Increasing the password strength is a solution to avoid dictionary attacks or to make brute force attacks infeasible. It is generally accepted that the length of the password determines the security it provides, however, it is not exactly true: the strength of the password is rather related to its entropy. As for example, a user that choose a password of 7 characters is said to provide between 16 and 28 bits of entropy.

Biometrics is often presented as a promising alternative[3]. The main advantage of biometrics is that it is usually more difficult to copy the biometric characteristics of an individual than most of other authentication methods such as passwords or tokens. Of course, many attacks exist and the goal of the researcher is to provide more robust systems. Fingerprints are the most used biometric modality even if it is quite easy to copy the fingerprint of an individual. In order to propose an alternative solution, recent research works propose to use the finger or hand veins as characteristics. The biometric technology is used for many applications such as logical and physical access control, electronic payments... There exist mainly three different types of biometric information: biological, behavioral and morphological information. Biological analysis exploits some information that are present for any alive mammal (DNA, blood). The behavioral analysis is specific to a human being and characterizes the way an individual makes some daily tasks (gait, talking) [4-5]. Last, morphological analysis uses some information on how we look like (for another individual or for a particular sensor) [6].

Among the different types of biometric systems, keystroke dynamics biometric systems seem to be an interesting solution for logical access control for many reasons:
- this biometric system does not necessitate any additional sensor,
- users acceptability is high as it is natural for everybody to type a password for authentication purposes,
- this kind of biometric system respects the privacy of users. Indeed, if the biometric template of an individual has been stolen, the user just has to change its password.

We propose to study in this paper how the use of this biometric system could be appropriated for the authentication of an individual who wants to access to a collaborative system. We consider security aspects (in terms of authentication errors), usability and collaborative environment constraints (essentially few required data from user to create his model and don't change his habits). The plan of the paper is given as follows. The next section gives an overview of the literature in the domain. Section three presents the results of a comparative study on many methods from the state of the art we realized. We discuss the ability of this authentication method to fit the requirements and constraints of this application (ease of use, speed, low organizational and technical impacts...). We conclude and give the different perspectives of this study.

## 2. STATE OF THE ART

Research about keystroke dynamics have begun more than twenty years ago. In [7], authors have made the first work to our knowledge on authentication with keystroke dynamics. They have shown that, like signature dynamics, keystroke typing can characterize a user. This study was done by a database consisting of the work of seven secretaries who have been asked to type three pages of text. We can easily see that this approach is not really an operational solution for collaborative applications (nobody would agree to type such a huge quantity of data), but many amelioration have been done since.

### 2.1. Main Approaches

The main scheme for keystroke dynamics is presented in Figure 1:
  (i) the enrollment (which consists of registering the user in the system) embeds the data capture, eventually some data filtering, the feature extraction, and the learning step;

(ii) the verification process realizes the data capture, the feature extraction and the comparison with the biometric model.

Depending on the studies, this scheme may be slightly modified. Some algorithms can also adapt the model of the user by adding the template created during the last successful authentication.

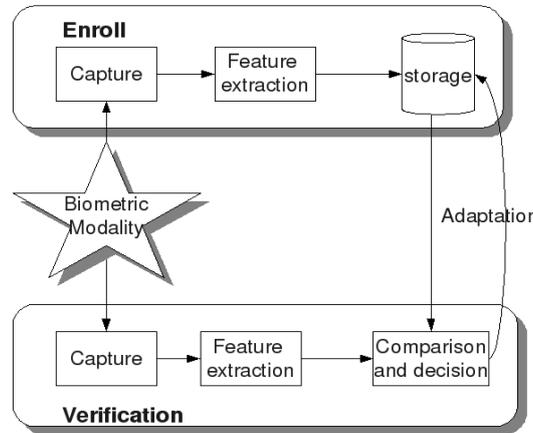

Figure 1. General Synopsis Of A Common Biometric System.

Two main families of keystroke dynamics systems exist. The first one is based on free texts. During the enrollment, a huge quantity of text have to be typed to create the user's model. For the verification, the procedure may be different because the user is asked to type different texts. The other family uses static texts : the same password is used for both enrollment and authentication.

In most studies, the quantity of information used to create the profile during the enrollment process is really huge. Systems using neural networks need hundred of data and most of statistical methods need at least twelve captures. This huge quantity of data is an obstacle for users.

## 2.2. Principles

In most articles of the literature, the following data are captured (see Figure 2): (i) the hold time of a key : which is the duration of key pressure (T3-T1) and (ii) the inter-key time : which is the delay between the release of a key and the pressure of the next one (T3-T2).

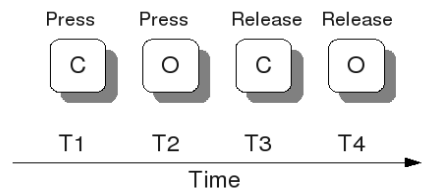

**Figure 2. Capture Information In Keystroke Dynamics Systems**

Other measures can also be captured [8] : the duration of password typing, the use of "invisible keys" (like BACKSPACE) in order to add entropy (or count the typing error rate of the user).

Data cleaning can be done to improve the system's performance. In [9], times superior to 500ms are ignored (they are considered as a result of a perturbation of the user). In [10], a reduction of the dimensions is done by using genetic algorithm associated to the Support Vector Machine algorithm.

Different features extraction methods exist. Sometimes, the raw data are used without any transformations. Other times, researchers prefer to work with n-graphs instead of di-graph (i.e., they use the association of more than two letters, three seems to be a good number as compromise). In [9] the extracted features are simple vectors of four dimension: (i) Manhattan distance of inter key time, (ii) Manhattan distance of hold time, (iii) Euclidean distance of inter key time, and (iv) Euclidean distance of hold time.

Depending on the study, different algorithms are used during the learning procedure. The most classical methods of model creation, in the literature, are the following: (a) computing of the mean and standard deviation of time vectors, (b) creating clusters with different algorithms (Support Vector Machine, K-nearest neighbor, K-means), (c) Bayesian classification, (d) neuronal networks.

During the verification process, the user is told to type the password. The resulting template of this input is compared with the biometric model of the claimed user and a distance score or a class is returned. The main procedures are: (a) minimal distance computing, (b) statistical methods, (c) verification of class matching, (d) time based discretization or (e) bioinformatics tools. The score is compared with a threshold in order to authenticate or not the user. So, the robustness of the biometric system depends a lot of this threshold configuration.

## 2.3. Existing Methods

Table 1 shows the estimated Error Equal Rate (corresponding to a compromise error that will be formally defined later in the paper) of several methods presented in the literature. As most articles do not provide value for the EER, an estimated value is presented in Table 1, we compute it by doing the average of errors of rejected genuine users and accepted impostors. This estimated error rate, for most of the articles, is lower than 10% and the best rates are obtained while using data-mining methods which need many data and don't fit our requirements.

**Table 1. Authors, Methods And Estimated EER Ordered By Year Of Publication**

| Article | Method | Estimated EER |
|---|---|---|
| Umphress and Williams (1985) [11] | statistical | 9.00% |
| Bleha et al (1990) [12] | bayes | 5.45% |
| Obaidat and Sadoun (1997) [13] | neuronal networks | 5.8% |
| Robinson et al (1998) [14] | statistical | 10,00% |
| Coltell et al (1999) [15] | statistical | 17.5% |
| Bergadano et al (2002) [16] | disorder mesure | 2,00% |
| Kacholia and Pandit (2003) | cauchy distribution | 2.9% |
| Guven and Sogukpinar (2003) [17] | cosine | 6.35% |
| Yu and Cho (2004) [18] | SVM | 3.14% |
| Rodrigues et al (2006) [19] | HMM | 3.6% |
| Clarke and Furnell (2006) [20] | neuronal networks | 5,00% |
| Filho and Freire (2006) [21] | HMM and statistical | 6,00% |
| Hwang et al (2006) [9] | SVM | 0.25% |
| Hoquet et al (2007) [22] | statistical and fusion | 5,00% |
| Revett et al. (2007) [23] | bioinformatic | 0.175% |

### 2.4. Discussion

Performance of systems can be improved by using multi-modality: the fusion of the scores from several methods is operated.

Because most systems use a very huge quantity of data during the enrollment step, those methods are not easily usable in a real life application such as collaborative systems. The user needs too much investments and will prefer keeping a classical password authentication solution. It is a real problem because keystroke dynamics is one of the less intrusive biometric system. That's why, in the next part, a comparative study is presented by keeping in mind the fact that the system has to be used in a collaborative system (only five vectors, or inputs, compose the model and the response is instantaneous).

Another way of improving the system's performances is to consider the fact that the user learns how to type the password all along his typing. That's why it's quite important to add the last template to the model.

## 3. COMPARATIVE STUDY

To have a good overview on the performances of the existing methods for an operational use, we have created an experimental protocol that aims to test several points: the capture process, the performance of existing algorithms and the perception of users about the system. It's aim is also to compare original performances in a specific environment with the performance in a more realistic environment. First, we present the experimental protocol, then the selected methods from the state of the art for this experiment, and then the results we obtained.

### 3.1. Experimental Protocol

#### 3.1.1. Test Population
We asked to sixteen individuals, during three sessions, to enroll themselves in our biometric system with the password "laboratoire greyc". During each session, the user was asked to type correctly five times the password (there is fifteen vector per user). The first five vectors are used for the enrollment process of the biometric system, while the others are used to test the system.

The individuals who took place to the protocol have a daily use of computer, they are familiar with a keyboard. The test population is composed of thirteen males and three females and their ages are between twenty-three and forty-five years old.

**3.1.2. Acquisition Procedure**
In order to take into account typing learning problems (modification of typing all along the time), the sessions took place during different days. The first session took place the 14th and 17th of November, and session 2 and 3 took place respectively the 18th and 19th of November.
The users could come when they wanted.

**3.1.3. Captured Data**
As presented in [22], the acquired biometric templates contain: (i) the time between two keys pressure (T2-T1) noted PP, (ii) the time between two keys release (T4-T3) noted RR, (iii) the time between one release and one pressure (T2-T3) noted RP, (iv) time between one pressure and one release (T3-T1) noted PR. In addition of these data, other were captured, they are presented in the performance evaluation section.

**3.1.4. Keystroke Dynamics Algorithms**
Five matching methods were tested. These methods are presented in the literature and seem to match the best of our requirements (few template used for enrollment, computationally simple).
The first tested method is presented in [12]. The value $v$ is the column test vector and $u$ is the column mean vector of enroll vectors :

$$score_1 = \frac{\overline{v-u}^t \overline{v-u}}{\|v\| . \|u\|} \quad (1)$$

The second tested method is presented in [22]. It is a statical method based on the average ($u$) and standard deviation vectors ($s$) computed with the enrollment vectors, with $v$ a test vector of $n$ dimensions:

$$score_2 = 1 - \frac{1}{n} \sum e^{\frac{-\overline{v_i - u_i}}{s_i}} \quad (2)$$

The third method was initially created for free text, but also seemed adapted for passwords [24]. It computes the Euclidean distance between the test vector with each enrolled vector and keeps the minimal one.

$$score_3 = min_{u \in enrollvectors} \sqrt{\sum \overline{u_i - v_i}^2} \quad (3)$$

Another score method consists in computing the square of the norm of the average enrolled vectors subtracted to the test vector [21].

$$score_4 = \|v - u\|^2 \quad (4)$$

The last method is based on the Euclidean distance between units vectors. The distance of unit test vector is compared with the distance of unit mean of the unit enroll vectors.

$$score_5 = \sqrt{\sum \overline{u_i - v_i}^2} \quad (5)$$

The first four methods are tested in an off-line mode (with registered data and without any user presence), while the fifth one is used in an on-line mode (while the user is using the application).

Different kinds of biometric templates have also been tested (the four vector time presented before plus a vector (V) consisting of their concatenation). The user changes his way of typing as long as he types it. In order to confirm that fact, the following methods have also been tested by : (i) keeping the five first vectors for enrollment (static mode), (ii) releasing the last enrollment vector and replacing it by the last tested one (adaptive mode), (iii) adding the last tested vector to the set of enrolled vectors (progressive mode). By using these variations, the number of different alternatives of each off-line method is 15 so, 60 different flavors are tested.

## 3.2. Performance Evaluation

It is necessary to analyze the performance of such a system to know if it is usable in a collaborative environment. Objective and subjective evaluations must be both realized.

### 3.2.1. Algorithms Performances
The efficiency of a biometric algorithm can be analyzed with the following measures : (i) the False Acceptation Rate (FAR), which is the proportion of authenticated impostors, (ii) the False Rejection Rate (FRR), which is the proportion of rejected genuine users, (iii) the Failure To Acquire rate (FTA), which is the proportion of biometric template that the system could not capture, (iv) the Error Equal Rate (EER), which the rate when the FAR is equal to the FRR, and (v) the ROC curves which plot FRR versus FAR for all the possible threshold values. It is a good representation of the performance of the algorithm.

### 3.2.2. Biometric Modality Performance
In addition to the biometric acquisition, other data have been captured : (a) the time to type the whole password, (b) the time to type all the passwords during the session, (c) the number of errors during the session (when the user makes a mistake while typing the password, he cannot correct it and has to type it again from scratch), and (d) subjective evaluation questionnaire during the last session.

### 3.2.3. User Perception
At the end of the last session, users were asked (I) to try to authenticate themselves three times in the system, (ii) to try to identify themselves and (iii) to answer to the following questions :

```
Q1: Is the verification fast? Yes, no
Q2: Is the system easy to use? Yes, no
Q3: Are you ready to use this system in the future? Yes, no, do not know
Q4: Do you feel confident in this system? Yes, no
Q5: Do you feel embarrassed when using this system? Yes, no, do not know
Q6: What is your general appreciation of the system? Very good, good, average, bad
```

The aim of this questionnaire is to give an idea of the user's perception after using the system. It gives us some information on its acceptance and if it could be used in real life (Where is the interest to do this kind of authentication in collaborative environment if no users agreed with the system ?). The questionnaire was presented after the steps of testing authentication and identification .

## 3.3. Experimental Results

Several types of results can be analyzed: the figures extracted during the whole protocol, the biometric algorithm performance and the answers of the subjective analysis.

### 3.3.1. Acquisition
Due to the learning typing effects, and different levels of user concentration during the sessions, the timing results are slightly different for the data of the first session and the whole data set. For most users, the minimum and average typing speed of the password have decreased (they ameliorate their way of typing), and the maximum time of typing has increased (due to the lack of concentration and perturbations from environment) for very few of them .

We can also see that the standard deviation of typing speed is more important with the whole data than with the data of one session. This can be interpreted by the problem of learning typing. That is why it is preferable to work with the last patterns and not all of them to avoid this effect. The users with a biggest increase of standard deviation (so, the less constant typing users), are the ones with an increase of maximum time.

The timings for the the sessions have also been compared. They are shown in the figure 3. The duration of the first session is more important than the two others, because it has consisted of : (i) quickly explain the project (most of them were already aware of what is related to), (ii) explain what the user has to do, (iii) type two times the password (to adapt himself to the keyboard , the chair and the password), (iv) add the user to the application, (v) type the five valid passwords. The others sessions have only consisted in (i) selecting the user, (ii) typing the five valid passwords. Except one user, the time for session 2 and 3 are quite similar.

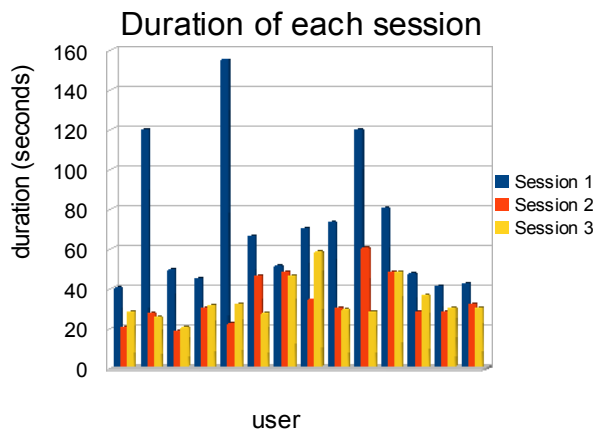

**Figure 3. Values On Durations Off Time Typing For Each User For All The Three Sessions**

The captured data are the keystroke timing of valid passwords (a typing error implies to type all the password again). The Figure 4 shows the FTA rate for each user (X axis represents the users and Y axis is the percentage of error during enrollment). The average FTA is about sixteen percent, which is a quite huge number. Users with the biggest FTA are: (i) the one who wanted to type too fast and did mistakes, (ii) the ones who are not concentrated or disturbed while typing, and (iii) others who do not feel very comfortable with the protocol.

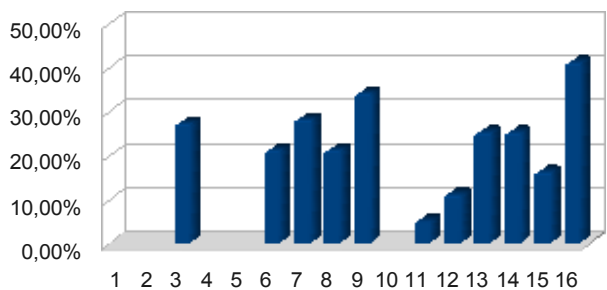

**Figure 4. FTA Of Each User For The Whole Protocol.**

### 3.3.2. On-line Performance

At the end of the last session, users has to authenticate themselves three times and identify themselves one time. The Figure 5 shows the results of this analysis.

64% of users were correctly identified in one try. 50% of users have been correctly authenticated during all of the three tests, 85% have been correctly authenticated at least two times, and 100% of users have been correctly authenticated at least one time. In a system allowing only three password tests before locking the account, all the users would have been authenticated.

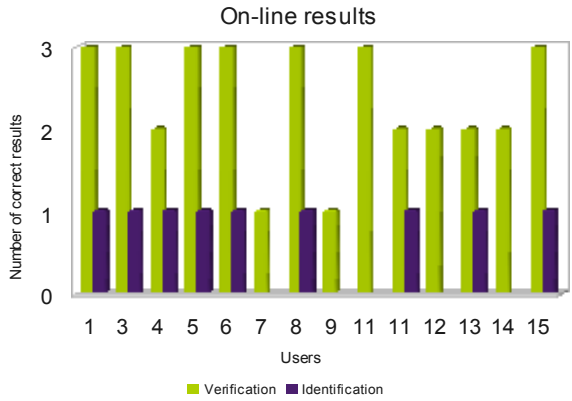

**Figure 5. Results Of The Three Authentications And The Identification Of The Users**

### 3.3.3. Off-line Performance

Even if 100% of the users are authenticated in at worst three tries, the performances of the algorithm described by equation (5), and used during the protocol, are far of being good. The Figure 6 presents the score distribution (depending on the threshold) and the ROC curve.

The first curve indicates that to obtain the EER, the threshold of the system must configured to 0.28, and the second curve shows that the EER is superior of 20%. A such bad rate can be explained by several reasons: the poor quality of the algorithm (this work was done before doing the state of the art), the quite small number of users in the database, and the fact of using an identical threshold for all users.

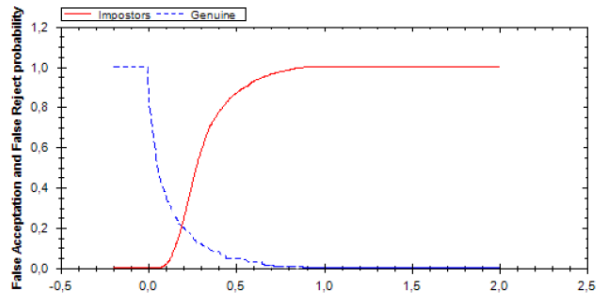
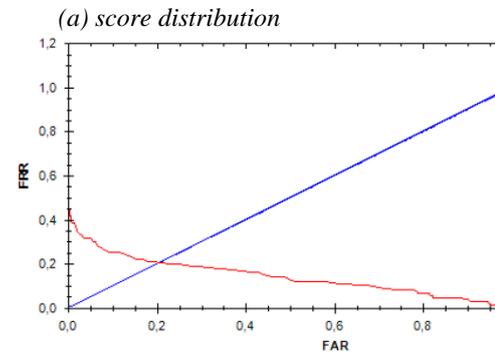

*(a) score distribution*

*(b) ROC curve*

**Figure 6. Off-line Performance Of The Method Used In An On-line Way**

Table 2 gives the EER, computed with the best parameters, of each tested method. The first column is the number of the tested method, the second column presents the best biometric template used, the third column shows which kind of enrollment method is used, the fourth column presents the EER when the threshold of decision is the same for all the users, and the last presents the EER when each user has his own optimized threshold. With our database the best method gives an EER value of 9.78% when using a global threshold and 4.28 % when using an individual threshold.

**Table 2. Best Performance Of Off-line Methods From The State Of The Art.**

| Method | Best template | Best kind | EER 1 | EER 2 |
|---|---|---|---|---|
| 1 | RP | progressive | 11,64% | 10,24% |
| 2 | V | adaptive | 9,78% | 4,28% |
| 3 | RP | adaptive | 14,05% | 11,09% |
| 4 | RR | progressive | 19,42% | 14,06% |

We can see, that it is important to take into account the evolution of typing of users (in the four cases, best methods are the progressive or adapting one) and using individual thresholds give better results.

### 3.3.4. Discussion

We can see, from the results of the Table 1 and Table 2, that those results do not improve the results of the previous researches. This is mainly because our work was done in a more constrained environment (only five vectors used for the enrollment), no filtering is done on the users (all the users with the required number of vectors are taken into account in the study, even the ones who decrease the results) or the data and all the users used the same password (so the number of patterns to test attacks for each users is much bigger than the number of patterns to test genuine authentication).

It is not an easy task to obtain enough good results which can be used in a collaborative environment. The few number of enrollment may be the main point of this performance decrease, that is why we will have to find other methods to improve the results while keeping this few number of enrollment.

Although users answered to the questionnaire after testing the worst algorithm of this study, their confidence and general appreciation in the system is good for more than 60% of them.

The number of participants of this study is really too small as well as the number of sessions. Following the practices exposed in [25] is a good solution.

## 4. CONCLUSIONS AND PERSPECTIVES

We studied the ability of keystroke dynamics authentication systems for their application to collaborative systems (collaborative systems need to authenticate there users). We can see through this study that, even with quite simple methods from the state of the art, the obtained results are almost correct (method 2, with less than 5% of error) but need yet to be improved. One-class Support Vector Machines with only five vectors per users for the training seems to give better results.

In [8], the authors present some useful and easy techniques in order to improve the quality of a keystroke dynamics system without modifying the algorithms: it is up to the users of the system to add breaks in their way of typing (they can be helped by the software with visual or sonore cues). Maybe these good practices of composing keystroke dynamic based passwords could be better accepted than the good practice of classical passwords.

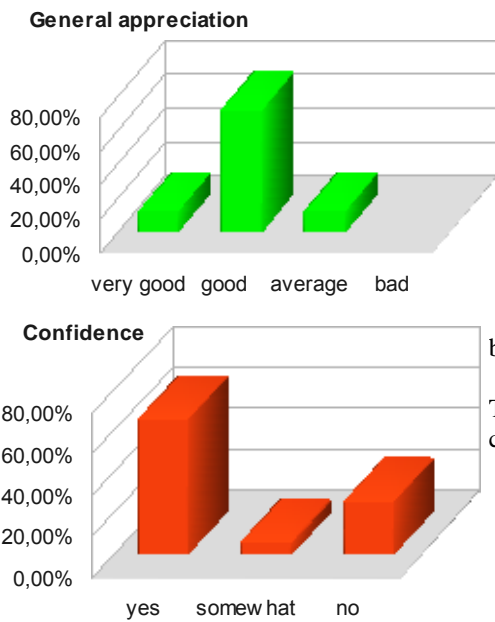

**Figure 7. Partial Results Of The Subjective Analysis**

In the tested systems, only well typed passwords are taken into account. This is a real problem, because with static password based authentication methods, the genuine user can correct himself his typing errors and being correctly authenticated. On the contrary, with our keystroke dynamic implementation, the system will force the user to type again the password in order to have a correct sized vector. The system and algorithms have to be modified to allow the use of backspace key to correct the password (because errors can be characteristic of the user).

There is a lack of data in the analysis of the subjective evaluation. It could be a good thing to use statistics tools to filter incoherent answers and get relations between the questions (with the help of Bayesian network ordecision trees). It gives better results to compute the robustness of all the algorithms. This could also give more information on the way of how to interpret this kind of curves.

## ACKNOWLEDGMENTS

The authors would like to thank the Basse-Normandie Region and the French Research Ministry for their financial support of this work.